%% file: main.tex
\documentclass[sigconf,nonacm]{acmart}

\AtBeginDocument{%
  \providecommand\BibTeX{{%
    \normalfont B\kern-0.5em{\scshape i\kern-0.25em b}\kern-0.8em\TeX}}}

\setcopyright{none}

\copyrightyear{}
\acmYear{}
\acmDOI{}

\input{package.tex}

\begin{document}

% \title{Slack Directly Annotated on Your Verilog: \\Fine-Grained RTL Timing Evaluation for Early Optimization}

% \vspace{-.2in}

\title{\vspace{-.4in}Dr.~RTL: Autonomous Agentic RTL Optimization through Tool-Grounded Self-Improvement}

% \affiliation{%

% \author{Wenji Fang}
% \email{wenjifang1@ust.hk}
% \affiliation{%
%   \institution{HKUST, HKUST(GZ)}
%   \city{}
%   \country{}
% }
\author{Wenji Fang, Yao Lu, Shang Liu, Jing Wang, Ziyan Guo,\\ Junxian He, Fengbin Tu, Zhiyao Xie}
\authornote{Corresponding Author (eezhiyao@ust.hk)}
\affiliation{%
  \institution{Hong Kong University of Science and Technology (HKUST)}
  \country{}
}

\pagestyle{plain}

\input{section/abstract.tex}

\maketitle

\input{section/1-intro.tex}

\input{section/2-preliminaries.tex}

\input{section/3-method.tex}

\input{section/4-experiments.tex}

\input{section/5-discussion.tex}
\input{section/6-conclusion}

\clearpage
\bibliographystyle{ACM-Reference-Format}
\bibliography{ref}

\end{document}

%% file: package.tex
\usepackage{hhline}
\usepackage{colortbl}
\usepackage{centernot}
\usepackage{pbox}
\usepackage{amsmath}
\usepackage{amsfonts}
\usepackage{url}
\usepackage{bm}
\usepackage{comment}
\usepackage{graphicx}
\usepackage{subfigure}
\usepackage{epstopdf}
\usepackage{multirow}
\usepackage{color}
\usepackage{tabu}
\usepackage{mdframed}
\usepackage{syntax}
\usepackage{fancyvrb}
\usepackage{cleveref}

\usepackage[ruled,linesnumbered,algo2e]{algorithm2e}
\usepackage[noend]{algpseudocode}
\usepackage{algorithmicx}
\usepackage{algpseudocode}
\usepackage{paralist}
\usepackage{stmaryrd}
\usepackage{tikz}
\usepackage[referable]{threeparttablex}
\usepackage{tabularx}
\usepackage{booktabs}
\usepackage{array}
\usepackage{fancyhdr}
\usepackage{float}
\usepackage{xspace}
\usepackage{pifont}
\usepackage{balance}
\usepackage{xcolor}% http://ctan.org/pkg/xcolor

\usepackage{tablefootnote}
\usepackage[normalem]{ulem}
\usepackage{bbding}

\usepackage{tikz}
\usepackage{xcolor}

\usepackage{graphicx}
\usepackage{adjustbox}

\usepackage{float}

\newfloat{figtab}{htb}{fgtb}
\makeatletter
  \newcommand\figcaption{\def\@captype{figure}\caption}
  \newcommand\tabcaption{\def\@captype{table}\caption}
\makeatother

\definecolor{princetonorange}{RGB}{255,143,0}

\definecolor{lightgreen}{RGB}{198, 224, 183}
\definecolor{lightred}{RGB}{240, 205, 176}
\definecolor{agentrtl}{HTML}{CDE09A}

\usepackage{geometry}

%% file: section/abstract.tex
\begin{abstract}

Recent advances in large language models (LLMs) have sparked growing interest in automatic RTL optimization for better performance, power, and area (PPA). However, existing methods are still far from realistic RTL optimization: their evaluation is often unrealistic, based on manually degraded, small-scale RTL designs and weak open-source toolchains; their methods are also limited, relying on coarse design-level feedback and simple pre-defined rewriting rules.
To address these limitations, we present Dr.~RTL, an agentic framework for RTL timing optimization in a realistic evaluation environment, with continual self-improvement through reusable optimization skills. 
We establish a realistic evaluation setting with more challenging RTL designs and an industrial EDA workflow. Within this setting, Dr.~RTL performs closed-loop optimization through a multi-agent framework for critical-path analysis, parallel RTL rewriting, and tool-based evaluation. We further introduce group-relative skill learning, which compares parallel RTL rewrites and distills the optimization experience into an interpretable skill library. Currently, this library contains 47 pattern--strategy entries for cross-design reuse to improve PPA and accelerate convergence, and it can continue evolving over time.
Evaluated on 20 real-world RTL designs, Dr.~RTL achieves average WNS/TNS improvements of 21\%/17\% with a 6\% area reduction over the industry-leading commercial synthesis tool.\looseness=-1
% This suggests that non-trivial RTL optimization opportunities remain even after strong commercial synthesis and can be systematically discovered through our Dr. RTL. \looseness=-1

% Experiments on 20 real-world RTL designs show that Dr.~RTL consistently improves over commercial synthesis and outperforms prior LLM-based baselines, achieving average WNS/TNS improvements of 21\%/17\% with a 6\% area reduction. These results show that meaningful RTL optimization opportunities remain beyond synthesis alone and can be systematically discovered through tool-grounded agentic optimization.\looseness=-1
\end{abstract}

%% file: section/1-intro.tex
\vspace{-.1in}
\section{Introduction}\label{sec:intro}

Register-transfer level (RTL) design is a critical abstraction in modern digital ICs, manually developed by human engineers to bridge architectural specification and downstream logic synthesis. 
The quality of RTL largely determines the optimization space available to the subsequent synthesis and physical design tools. Modern RTL designs often incorporate deep pipelines, wide arithmetic datapaths, and tightly coupled control–data interactions~\cite{vahid2010digital}, all of which make timing closure more challenging.
Bottlenecks introduced at the RTL stage often propagate through the design flow, leading to repeated RTL revisions, conservative timing constraints, and suboptimal PPA trade-offs. Consequently, improving RTL quality before synthesis remains an important yet challenging task.\looseness=-1

\begin{figure}[!t]
  \centering
  % \vspace{-.05in}
  \includegraphics[width=1\linewidth]{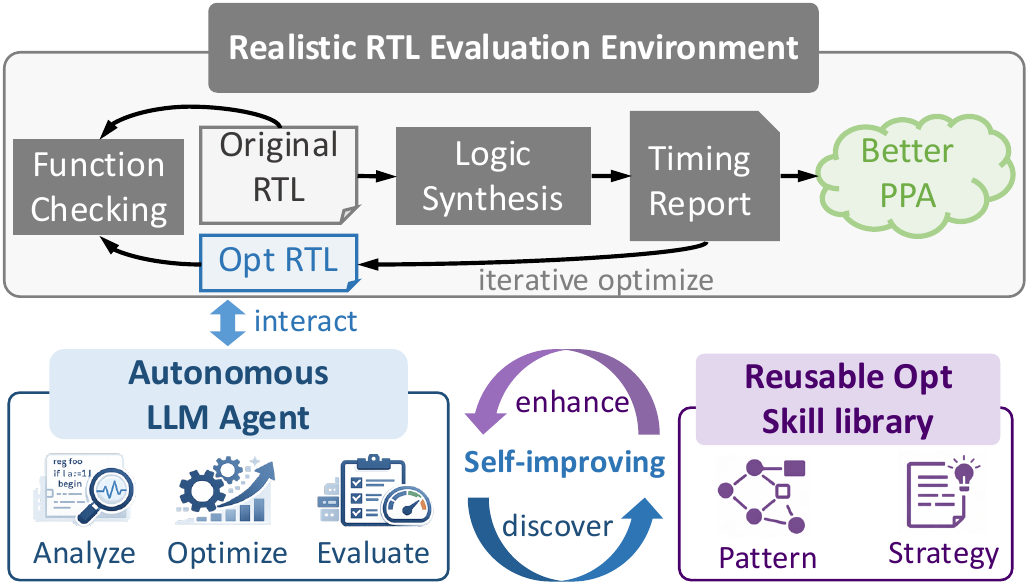}
  \vspace{-.25in}
  \caption{Dr.~RTL iteratively optimizes RTL PPA via closed-loop interaction with industrial EDA tools, while distilling trajectories into reusable skills for self-improvement.\looseness=-1}
  \label{fig:motivation}
  \vspace{-.2in}
\end{figure}

\textbf{Automatic synthesis optimization vs. manual RTL optimization.}
Synthesis and RTL optimization operate at different abstraction levels and complement each other. 
Synthesis improves PPA by transforming a fixed RTL into an optimized gate-level implementation through Boolean rewriting and technology mapping~\cite{brayton2006scalable, hassoun2012logic, ziegler2016synthesis, chowdhury2024retrieval, chen2024syn, yin2025boole, liu2026survey}, but it remains constrained by the original RTL structure and cannot change higher-level design choices such as datapath organization or pipelining. Consequently, further PPA improvement often requires iterative manual RTL rewriting guided by synthesis feedback: engineers identify critical paths with timing reports, revise the RTL, and re-run synthesis to evaluate each change, as illustrated in~\Cref{fig:motivation}.
However, this process remains labor-intensive, time-consuming, and heavily dependent on implicit human expertise, making it difficult to automate systematically. More fundamentally, the RTL design space is vast, effectively infinite for practical purposes, while human-driven optimization typically evaluates only one candidate design per iteration, exploring only a negligible fraction of the possible optimization space.

% Synthesis improves PPA by transforming a fixed RTL into an optimized gate-level implementation through Boolean rewriting and technology mapping~\cite{brayton2006scalable, hassoun2012logic, ziegler2016synthesis, chowdhury2024retrieval, chen2024syn, yin2025boole, liu2026survey}. However, it is constrained by the original RTL structure and cannot modify high-level design choices such as datapath organization or pipelining, limiting its optimization potential.\looseness=-1

% In practice, further PPA gains often require iterative manual RTL rewriting guided by synthesis feedback. As shown in~\Cref{fig:motivation}, traditionally, engineers analyze timing reports to identify critical paths and manually modify the RTL to fix bottlenecks, repeatedly invoking synthesis to evaluate each change. This process is labor-intensive and time-consuming, and relies heavily on implicit human expertise, with little systematic knowledge for automation.\looseness=-1

\begin{table*}[!t]
\vspace{-.4in}
\caption{Comparison with existing LLM-based RTL optimization benchmarks and methods.\looseness=-1}%applied in optimization.}
\vspace{-.15in}
\resizebox{1\textwidth}{!}{ 

\begin{tabular}{c||cccc|cc|c|c} \toprule
                                      & \multicolumn{4}{c|}{\textbf{RTL Dataset}}                                                                                                                                      & \multicolumn{2}{c|}{\textbf{Tool Use}}                                                      & \multirow{2}{*}{\textbf{Planning}}                                                & \multirow{2}{*}{\textbf{Memory}}                                    \\ \cmidrule{2-7}
\textbf{Work}                         & \small{LoC$^\star$}                            & \small{NoM$^\star$}                         & \small{Complexity}              & \small{Opt Input}                                                                            & \small{EDA Tool}                  & \small{PPA Feedback} &                                                                                   &                                                                     \\ \midrule
RTLRewriter~\cite{yao2024rtlrewriter} & [8, 86, 1275] & [1, 1, 14] & \multirow{4}{*}{Module} & \multirow{5}{*}{\begin{tabular}[c]{@{}c@{}}Manual-degraded \\      RTL\end{tabular}} & Yosys + Simulation               & \multirow{5}{*}{\begin{tabular}[c]{@{}c@{}c@{}}Coarse-grained \\  design PPA only\end{tabular}  }                                      & \multirow{4}{*}{Single LLM}                                                       & \multirow{5}{*}{\begin{tabular}[c]{@{}c@{}c@{}}Pre-defined \\  rewriting rules\end{tabular}  }                                                \\
SymRTLo~\cite{wang2025symrtlo}        & \multicolumn{2}{c}{Same as~\cite{yao2024rtlrewriter}}                             &                         &                                                                                      & Yosys/DC+ Simulation &                                                                &                                                                                   &                                                                     \\
RTL-OPT~\cite{lu2026new}              & [8, 31, 135]                   & [1, 1, 1]                   &                         &                                                                                      & DC + CEC$^\dagger$                           &                                                                 &                                                                                   &                                                                     \\
POET~\cite{ping2026poet}              & \multicolumn{2}{c}{Same as~\cite{lu2026new}}                                       &                        &                                                                                      & Yosys + Simulation                           &                                                                 &                                                                                   &                                                                     \\
CODMAS~\cite{chang2026codmas}              & \multicolumn{3}{c}{Not publicly accessible}                                                     &                                                                                      & Yosys + Simulation                           &                                                                 &      Multi-LLM                                                                             &                                                                     \\ \midrule \midrule
\cellcolor{agentrtl}\textbf{Dr. RTL (ours)}                       & \cellcolor{agentrtl}\textbf{[128, 812, 4615]}               & \cellcolor{agentrtl}\textbf{[1, 3, 7]}                   & \cellcolor{agentrtl}\textbf{IP/Design}               & \begin{tabular}[c]{@{}c@{}}\cellcolor{agentrtl}\textbf{Original human-} \\      \cellcolor{agentrtl}\textbf{written RTL}\end{tabular}                    & \cellcolor{agentrtl}\textbf{DC + SEC$^\dagger$}                  & \begin{tabular}[c]{@{}c@{}}\cellcolor{agentrtl}\textbf{Fine-grained}\\      \cellcolor{agentrtl}\textbf{critical path}\end{tabular}  & \begin{tabular}[c]{@{}c@{}}\cellcolor{agentrtl}\textbf{Orchestrator} \\      + \cellcolor{agentrtl}\textbf{Sub-agents}\end{tabular} & \begin{tabular}[c]{@{}c@{}}\cellcolor{agentrtl}\textbf{Skill self-}\\      \cellcolor{agentrtl}\textbf{improving}\end{tabular} \\ \bottomrule
\end{tabular}

}
\begin{tablenotes}\footnotesize
\item $^\star$ We report the [minimum, average, maximum] lines of code (LoC) and number of modules (NoM) for each dataset. Our dataset is around 10$\times$ larger than existing ones.
\item $^\dagger$ Prior work relies on combinational equivalence checking (CEC), which cannot verify sequential changes. Our method employs sequential equivalence checking (SEC), supporting both combinational and sequential optimization scenarios.\looseness=-1
\end{tablenotes}
\label{tbl:baseline}
\vspace{-.1in}
\end{table*}

\textbf{LLM for RTL Design.}
To reduce this manual effort, recent advances in large language models (LLMs) have sparked growing interest in RTL design automation, as summarized in recent surveys~\cite{chen2024dawn, pan2025survey, fang2025survey, he2025large, ravindran2025survey, zang2025dawn}. Most prior works~\cite{liu2024rtlcoder, zhao2024codev, pei2024betterv, liu2024openllm, pinckney2025comprehensive, delorenzo2024make, thorat2025llm, tasnia2025veriopt, thorat2023advanced, min2025revolution, akyash2025rtl++, yu2025spec2rtl, hsin2026evolve, allam2024rtl, auto-v-coder, sami2024aivril, zhao2025mage, deng2025scalertl, deng2026ace, zhang2026rtlseek, chen2026incrertl} focus on generating RTL from natural language specifications, with functional correctness as the primary objective and PPA only weakly considered. They are also typically evaluated on small-scale benchmarks such as VerilogEval~\cite{liu2023verilogeval} and RTLLM~\cite{lu2024rtllm}, which fall short of the complexity and critical-path behavior of real-world RTL designs. More recently, several studies~\cite{yao2024rtlrewriter, wang2025symrtlo, xu2025rethinking, lu2026new, chang2026codmas, ping2026poet} start to explore \emph{RTL optimization}, aiming to rewrite existing RTL for better PPA while preserving functionality, as summarized in~\Cref{tbl:baseline}.

\textbf{Limitations of existing LLM-based RTL optimization.}
Despite recent progress, existing approaches remain far from addressing realistic RTL optimization in both evaluation and methodology: 

From an \textit{evaluation} perspective, they suffer from three key issues:
\emph{(1) Weak starting RTL.} Prior works~\cite{yao2024rtlrewriter, wang2025symrtlo, xu2025rethinking, lu2026new, chang2026codmas, ping2026poet} evaluate on manually degraded RTL with injected degradation or redundancies.
This makes the task easier and less realistic, reducing it to PPA repair rather than true optimization of well-written RTL. 
% As a result, such settings do not fully demonstrate the ability to discover non-trivial optimization opportunities.
\emph{(2) Unrealistic toolchains.} Most studies~\cite{yao2024rtlrewriter, wang2025symrtlo, xu2025rethinking, chang2026codmas, ping2026poet} rely on open-source synthesis tools that are much weaker than commercial ones, so some reported gains may already be absorbed by commercial tools and may reflect tool limitations rather than true RTL optimization.
\emph{(3) Very limited scale.} Existing evaluations~\cite{yao2024rtlrewriter, wang2025symrtlo, xu2025rethinking, lu2026new, chang2026codmas, ping2026poet} focus on tiny modules and fail to reflect the scale and complexity of industrial RTL designs.\looseness=-1

From a \textit{methodology} perspective, existing methods are limited by two factors. First, they rely mainly on LLM-based code inspection, without fine-grained EDA feedback to guide critical-path optimization. Second, they depend on pre-defined basic rewriting rules from textbooks~\cite{yao2024rtlrewriter, wang2025symrtlo}, which are often ineffective under modern synthesis tools and restrict the discovery of non-trivial optimization strategies.
These gaps motivate the following research questions:\looseness=-1
\begin{enumerate}
    
    \item [\textbf{Q1.}] \textbf{Evaluation:} How should RTL optimization be evaluated to reflect the realistic industrial scenario?
    \item [\textbf{Q2.}] \textbf{Method:} How can we perform effective RTL optimization and discover reusable optimization knowledge?
    \item [\textbf{Q3.}] \textbf{Optimization Gap:} What optimization opportunities cannot be handled by industrial synthesis and therefore require RTL optimization?
    
\end{enumerate}

To address these questions, we propose \textbf{Dr.~RTL}\footnote{The RTL dataset, evaluation environment, agentic implementation, and learned optimization skills are open-sourced at https://github.com/hkust-zhiyao/Dr_RTL.}, a tool-grounded agentic RTL timing optimization framework with self-improving capability.
As shown in~\Cref{fig:motivation}, built on a realistic evaluation setting (\textbf{A1}), Dr.~RTL performs closed-loop optimization (\textbf{A2}) through iterative interaction with EDA tools while continually distilling reusable optimization knowledge into a skill library (\textbf{A3}). 
\looseness=-1

We first answer Q1: \textbf{A1. Realistic industrial evaluation.}
We establish a realistic evaluation environment for RTL timing optimization with 20 larger-scale, human-written designs. Compared with prior works, it improves over three aspects:
\textit{1) Realistic starting RTL,} starting from human-written RTL instead of manually degraded inputs, reflecting true optimization scenario;
\textit{2) Industrial-standard toolchain,} using commercial synthesis for strong optimization and sequential equivalence checking (SEC) to ensure both combinational and sequential equivalence, ensuring high-fidelity PPA evaluation; and
\textit{3) Larger design scale,} evaluating designs that are around 10$\times$ larger and more complex than prior small-scale designs, better capturing the hierarchy and critical-path complexity of real RTL.\looseness=-1

Under this evaluation, we answer Q2 with \textbf{A2. Tool-grounded self-improving agentic RTL optimization.} We propose two coupled techniques for effective RTL optimization and continual self-improvement: \textit{(1) Agentic optimization:} it formulates RTL optimization as a tool-grounded search problem over equivalent RTL transformations, using a multi-agent framework to iteratively perform timing analysis, RTL candidate rewriting, and tool-based evaluation. This enables adaptive exploration and exploitation over a large optimization space guided by fine-grained critical-path feedback.
\textit{(2) Group-relative skill learning:} Dr.~RTL compares parallel rewritten RTL candidates from the same parent RTL and distills their optimization trajectories into reusable, human-interpretable skills, enabling continual test-time self-improvement without supervision or parameter tuning.\looseness=-1

% Under this evaluation environment, we answer Q2 with \textbf{A2. Tool-grounded self-improving agentic RTL optimization.} RTL optimization is challenging because the transformation space is large and discrete, while all modifications must preserve functional equivalence. Dr.~RTL formulates this task as a tool-grounded search problem over equivalent RTL transformations. It employs a multi-agent framework that iteratively performs timing analysis, RTL candidate rewriting, and tool-based evaluation, enabling adaptive optimization guided by fine-grained critical path feedback. The framework follows an exploration--exploitation paradigm: diverse strategies are explored in parallel, while the best SEC-passing candidate is promoted across iterations for further refinement. Dr.~RTL further introduces group-relative skill learning, which compares parallel candidates derived from the same parent RTL and distills their trajectories into reusable, human-interpretable skills. By comparing candidates under the same optimization context, it enables continual test-time self-improvement without the need for supervision or parameter tuning.

We further answer Q3 with \textbf{A3. Reusable RTL optimization knowledge.} The learned skills capture three forms of reusable knowledge, enabling cross-design reuse for better PPA, fewer invalid transformations, and faster convergence. Specifically, \textit{(1) Timing bottleneck patterns:} realistic RTL timing bottlenecks are often recurrent, arising from structural patterns such as deep decode logic, wide comparisons, and mux-heavy selection, which reveal optimization opportunities beyond synthesis alone; \textit{(2) Effective strategies:} for these patterns, effective strategies consistently emerge, including pre-computation, decomposition, and selective register insertion; and \textit{(3) Invalid strategies:} many naive transformations are either absorbed by synthesis or violate equivalence, such as manual rebalancing of optimized logic or moving control updates across registers. In our current implementation, Dr.~RTL extracts 47 pattern--strategy entries and releases them as a public skill library, which can be further expanded by applying the framework to more designs and incorporating expert knowledge from RTL designers.

Together, Dr.~RTL takes a step toward practical agentic RTL optimization that improves at test time without requiring LLM fine-tuning. More broadly, it points to a new paradigm of \textit{agentic design automation}, in which humans define the task, VLSI workflow, and constrained EDA feedback, while the agent autonomously explores effective solutions and accumulates reusable knowledge within this environment for self-improvement.

The contributions of this work are summarized as follows:
\begin{itemize}
    \item \textbf{Problem.} We identify RTL timing optimization as a practical and underexplored problem, where strong human-written RTL must be improved beyond synthesis rather than merely repaired from degraded inputs.
    \item \textbf{Evaluation.} We establish a realistic evaluation for RTL optimization, featuring challenging human-written RTL, commercial synthesis, and sequential formal verification.\looseness=-1
    \item \textbf{Method.} We develop Dr.~RTL, a tool-grounded closed-loop agentic RTL optimization framework, together with group-relative skill learning for continual self-improvement.
    \item \textbf{Knowledge.} We extract explicit RTL optimization knowledge, consisting of 47 pattern--strategy pairs, for cross-design reuse to improve PPA and accelerate convergence.\looseness=-1
    \item \textbf{Result.} Across 20 diverse RTL designs, Dr.~RTL achieves average WNS/TNS improvements of 21\%/17\% while reducing area by 6\% over commercial synthesis, demonstrating Pareto optimization beyond simple timing--area trade-offs.\looseness=-1
\end{itemize}

% The contributions of this work are summarized as follows.
% \begin{itemize}
%     \item We identify RTL timing optimization as a practical and underexplored problem, where strong human-written RTL must be improved beyond synthesis rather than merely repaired from degraded inputs.
%     \item We establish a realistic industrial evaluation setting
%     \item develop Dr.~RTL, a tool-grounded closed-loop optimization framework that performs adaptive exploration-exploitation over a large and discrete RTL transformation space, enabling RTL optimization beyond synthesis. introduce group-relative skill learning to distill recurring timing bottlenecks and solutions
%     \item We  into a reusable skill library of effective pattern--strategy pairs and avoid strategies. The skills are reused across designs to improve PPA, reduce invalid transformations, and accelerate convergence.\looseness=-1
%     \item Across 20 diverse RTL designs, Dr.~RTL achieves average WNS/TNS improvements of 21\%/17\% with a 6\% area reduction, demonstrating generalizable Pareto optimization beyond simple timing--area trade-offs.
% \end{itemize}

\begin{figure*}[!t]
  \centering
  \vspace{-.4in}
  \includegraphics[width=1\linewidth]{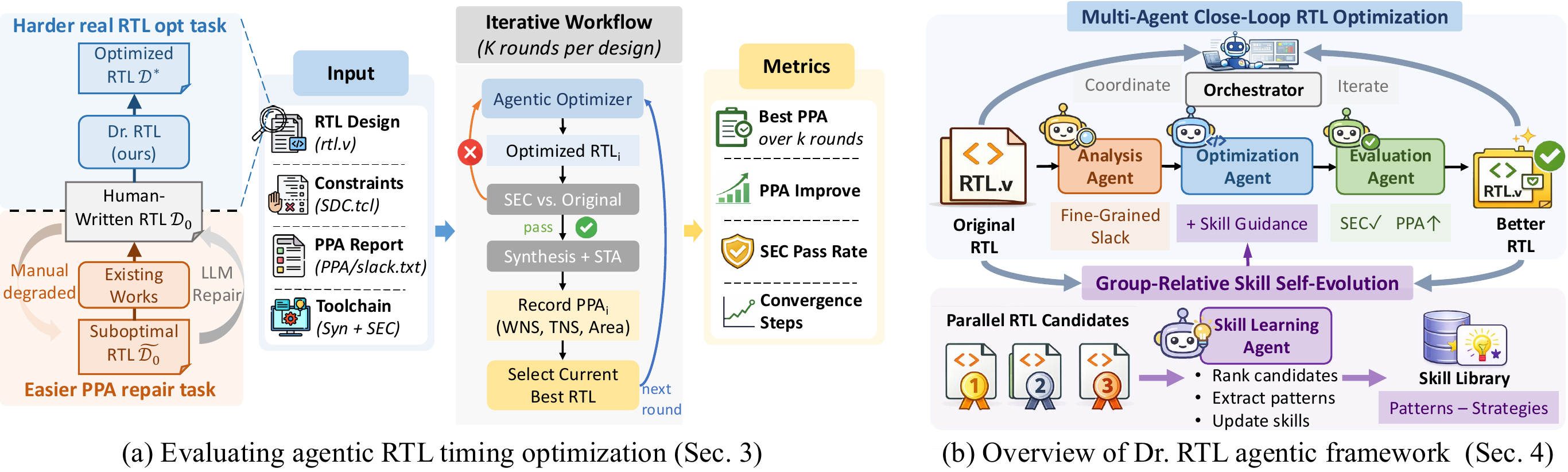}
  \vspace{-.32in}
  \caption{Proposed industrial-standard RTL optimization evaluation and overview of Dr.~RTL for agentic RTL optimization.} 
  \label{fig:overview}
  \vspace{-.1in}
\end{figure*}

%% file: section/2-preliminaries.tex
\section{Problem Formulation}\label{sec:prob}

We formulate RTL timing optimization as an \emph{agentic optimization problem}, where the LLM agent iteratively interacts with a commercial EDA environment. The synthesis and verification flow evaluates candidate RTL designs, and the agent proposes transformations based on the resulting feedback. Formally, given an initial RTL design $\mathcal{D}_0$, the goal is to obtain an optimized design $\mathcal{D}^*$ that improves post-synthesis PPA while preserving functional equivalence. We assume fixed synthesis settings and preserve the original micro-architecture, including pipeline latency. Under this constraint, we allow both combinational transformations and latency-preserving sequential restructuring, such as retiming-like register redistribution or duplication.
The optimization objective is defined as:
\begin{equation}
\min_{\mathcal{D}} f(\text{WNS}, \text{TNS}, \text{Area})
\quad \text{s.t.} \quad
\mathcal{D} \equiv \mathcal{D}_0 .
\end{equation}

For PPA evaluation, we focus on timing optimization while controlling area as a trade-off. Power is excluded from the current objective because accurate power estimation is workload-dependent and cannot be reliably assessed through static evaluation alone.\looseness=-1

%% file: section/3-method.tex
\section{Agentic RTL Optimization Evaluation Environment}
\label{sec:evaluation}

\Cref{fig:overview}(a) illustrates our industrial-standard evaluation environment for agentic RTL timing optimization. Compared with prior benchmarks~\cite{yao2024rtlrewriter, lu2026new, chang2026codmas}, our setup improves over three dimensions: input designs, toolchain, and evaluation metrics.

\textbf{(1) Input designs.}
We evaluate on original human-written RTL designs $\mathcal{D}_0$ collected from diverse open-source projects, rather than manually degraded RTL $\tilde{\mathcal{D}}_0$ constructed by injecting artificial degradations or redundancies into a reference design $\mathcal{D}_0$. As summarized in~\Cref{tbl:baseline}, prior benchmarks are typically limited to small single-module designs, often with only tens of lines of code. In contrast, our dataset contains substantially larger and more complex designs, with 812 lines of code on average, deeper module hierarchies, and diverse design types and coding styles. This setting better reflects realistic RTL optimization to improve already strong human-written RTL.\looseness=-1

% \textbf{(1) Realistic human-written RTL as input.}
% As shown in~\Cref{fig:agent}(a), prior works all adopt a weaker input by starting from degraded RTL $\tilde{\mathcal{D}}_0$, obtained by injecting artificial issues or redundancies into a reference design $\mathcal{D}_0$. Their RTL designs are also limited to very small-scale single-module designs, typically only tens of lines of code, as summarized in~\Cref{tbl:baseline}.

% In our work, we directly start from the original human-written RTL $\mathcal{D}_0$ collected from diverse open sources and seek further improvement.
% Our RTL designs are substantially larger (812 LoC on average), with deeper module hierarchies and various design types and coding styles. This setting better reflects practical scenarios, where optimization aims to improve already high-quality RTL.\looseness=-1

\textbf{(2) Commercial synthesis with complete formal verification.}
We adopt an industrial-standard EDA workflow that combines commercial logic synthesis with sequential formal equivalence checking (SEC). Commercial synthesis provides stronger and more realistic optimization than open-source tools such as Yosys~\cite{yao2024rtlrewriter, wang2025symrtlo, ping2026poet, chang2026codmas}, reducing the chance that simple RTL rewrites appear effective only because the synthesis is weak.
SEC is used to ensure equivalence after both combinational and sequential transformations. This is stronger than simulation-based checking~\cite{yao2024rtlrewriter, wang2025symrtlo, ping2026poet, chang2026codmas}, which is incomplete, and combinational equivalence checking~\cite{lu2026new}, which cannot support sequential rewrites. Together, this workflow enables reliable evaluation of RTL optimization under realistic industrial constraints.
% As shown in~\Cref{tbl:baseline}, prior works lack realistic and reliable evaluation. Most approaches~\cite{yao2024rtlrewriter, wang2025symrtlo, ping2026poet, chang2026codmas} rely on open-source tools such as Yosys, whose optimization capability is significantly weaker than commercial tools and may not reflect industrial performance. For verification, they typically use testbench-based simulation~\cite{yao2024rtlrewriter, wang2025symrtlo, ping2026poet, chang2026codmas}, which is labor-intensive and cannot guarantee full coverage. Some works adopt combinational equivalence checking~\cite{lu2026new}, but this fails to support sequential transformations, thereby restricting the optimization space.

% In contrast, we adopt a full industrial-standard EDA workflow that integrates commercial synthesis with sequential formal equivalence checking (SEC). This supports reliable validation of both combinational and latency-preserving sequential transformations.\looseness=-1

\textbf{(3) Agentic optimization metrics.}
To evaluate iterative agentic optimization, we report not only final PPA but also iteration metrics. Specifically, we measure: 1) \emph{best PPA} over all iterations, reflecting the peak optimization result; 2) \emph{PPA improvement} relative to the initial design $\mathcal{D}_0$; 3) \emph{SEC pass rate}, reflecting the validity rate of proposed transformations; and 4) \emph{convergence steps}, measuring how quickly the optimization stabilizes. Together, these metrics characterize both optimization quality and agentic behavior.
% Existing works~\cite{yao2024rtlrewriter, lu2026new, wang2025symrtlo, ping2026poet, chang2026codmas} focus only on final best PPA metrics, which are insufficient to evaluate the iterative agentic optimization. Such metrics fail to capture key properties including robustness, exploration efficiency, and convergence behavior.

% We introduce a set of \emph{agentic evaluation metrics} tailored for iterative optimization. Specifically, we measure:
% 1) \emph{best PPA} over all iterations, reflecting peak performance;
% 2) \emph{PPA improvement} relative to the initial design $\mathcal{D}_0$;
% 3) \emph{SEC pass rate}, indicating transformation success rate; and
% 4) \emph{convergence steps}, capturing optimization efficiency.
% Together, these metrics provide a comprehensive evaluation of agentic RTL optimization.

\begin{figure*}[!t]
  \centering
  \vspace{-.4in}
  \includegraphics[width=1\linewidth]{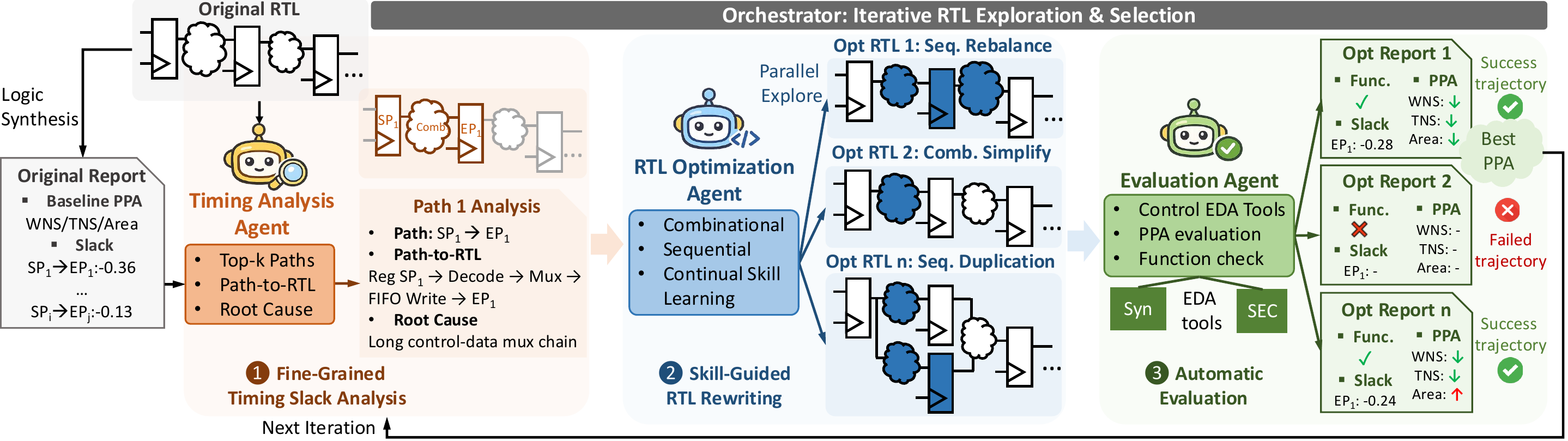}
  \vspace{-.3in}
  \caption{Multi-agent framework of Dr.~RTL. Dr.~RTL coordinates timing analysis, RTL optimization, and evaluation agents in a closed loop, using industrial EDA feedback to localize critical paths, explore RTL candidates in parallel, and promote the best SEC-passing design for iterative PPA improvement.}
  \label{fig:agent}
  \vspace{-.15in}
\end{figure*}

\section{Dr. RTL Methodology}\label{sec:method}

\Cref{fig:overview}(b) shows the overall architecture of Dr.~RTL. The framework combines two tightly coupled aspects: (1) a closed-loop optimization framework that iteratively performs timing analysis, RTL transformation, and tool-based evaluation under industrial EDA feedback via three specialized agents, as illustrated in~\Cref{sec:method1}, and (2) a learning mechanism based on group-relative skill learning, which compares parallel transformations under the same optimization context and distills effective experience into reusable pattern--strategy skills, as detailed in~\Cref{sec:method2}. Together, they enable continual improvement in both optimization quality and efficiency. We describe them below.

% To improve optimization over time, we propose a group-relative skill self-evolution mechanism. At each iteration, the generated candidate designs are evaluated and ranked based on their PPA quality. Their relative performance provides a structured signal for identifying effective transformations. A skill learning agent extracts reusable optimization patterns from high-quality candidates and stores them in a skill library $\mathcal{S}$, which is continuously updated to guide future RTL modifications.

\subsection{Agent Design and Responsibilities}\label{sec:method1}

\Cref{fig:agent} illustrates our orchestrator--agent framework. The orchestrator coordinates three specialized agents: (1) a timing analysis agent that identifies critical-path bottlenecks from post-synthesis timing reports; (2) an RTL optimization agent that rewrites $N$ candidate designs in parallel through equivalent transformations guided by the timing analysis; and (3) an evaluation agent that runs synthesis and equivalence checking to return PPA and correctness feedback for the next round of analysis.

% We decompose the workflow in this way because RTL optimization is heterogeneous and tool-driven, making a single monolithic prompt unreliable. Here, the \emph{workflow} refers to the fixed optimization loop of timing analysis, RTL transformation, and tool-based evaluation, whereas the \emph{agentic framework} specifies how specialized agents coordinate within and across these stages. This coordination is managed by the orchestrator through a shared JSON state that stores analysis results, candidate edits, evaluation feedback, and iteration history. In this way, the framework aligns each stage with industrial tool boundaries while enabling structured communication among agents.

% We use this decomposition because RTL optimization is heterogeneous and tool-driven, making a single monolithic prompt unreliable. Here, the \emph{workflow} refers to the fixed optimization stages, including timing analysis, RTL transformation, and evaluation, whereas the \emph{agentic framework} defines how specialized agents cooperate within and across these stages. 

This decomposition provides three benefits: (1) \emph{task specialization}, where each agent focuses on a well-defined role aligned with the industrial workflow, improving reliability and interpretability; (2) \emph{parallel exploration}, where parallel candidate generation and selection turn optimization into a tool-guided search over a non-convex design space; and (3) \emph{modular learning}, where the decomposition enables optimization behaviors to be attributed, analyzed, and reused. Overall, this design aligns agent roles with tool boundaries, leverages feedback, and balances exploration with refinement. We describe each agent below. The empirical benefit of the decomposition is shown in~\Cref{sec:abl}.

\subsubsection{Orchestrator.}

The orchestrator has three responsibilities: executing the optimization loop, selecting among parallel candidates across iterations, and maintaining trajectory logs for subsequent skill learning. Starting from an initial design $\mathcal{D}_0$, the system evolves over $t = 0,1,\dots,K$ following
\begin{equation}
\mathcal{D}_t \rightarrow \{\mathcal{D}_t^{(i)}\}_{i=1}^N \rightarrow \mathcal{D}_{t+1},
\end{equation}
where agents generate and evaluate $N$ candidate designs in parallel based on EDA tool feedback.

To coordinate the framework, the orchestrator maintains a shared JSON state that serves as the communication interface across agents. This state records timing analysis results, candidate RTL edits, evaluation outcomes, and iteration history, allowing each agent to consume structured outputs from previous stages and write back its own results. The same structured logs are also used to maintain optimization trajectories for subsequent skill learning. We provide a detailed illustration in~\Cref{sec:method2}.

To evaluate the PPA results, at each iteration, a scalar score is computed for each SEC-pass candidate:
\begin{equation}
\textit{Score}_i = \alpha\,\text{WNS}_i^{\text{norm}} + \beta\,\text{TNS}_i^{\text{norm}} + \gamma\,\text{Area}_i^{\text{norm}} + \text{penalty}_i,
\label{eq:score}
\end{equation}
where
$
\text{PPA}_i^{\text{norm}} = \frac{\text{PPA}_i - \text{PPA}_{\text{baseline}}}{\text{PPA}_{\text{baseline}}}
$, 
and $\alpha, \beta, \gamma \in [0,1]$ are weighting factors\footnote{In our experiments, we set $\alpha=0.5$, $\beta=0.35$, and $\gamma=0.15$. We set $\text{penalty}_i=0.5$ if $\text{Area}_i^{\text{norm}} > 0.1$, and $0$ otherwise, to discourage excessive area overhead. Different settings can be used to target different PPA trade-offs.}. Lower scores indicate better timing–area trade-offs. For WNS and TNS, improvement means moving closer to zero.
The next starting design is chosen as the best SEC-passing candidate:
\begin{equation}
\mathcal{D}_{t+1} = \arg \min_{\mathcal{D}_t^{(i)} } \textit{Score}_i, \quad \text{s.t.} \quad \text{SEC}_i = 1,
\end{equation}
where $\text{SEC}_i \in \{0,1\}$ indicates functional correctness.

% \subsubsection{Orchestrator.}

% The orchestrator is responsible for global decision-making across iterations. At each iteration $t$, it aggregates evaluation results from candidate designs and selects the next reference design based on both performance and correctness.

% The system operates in an iterative optimization loop. Starting from an initial design $\mathcal{D}_0$, it proceeds over $t = 0,1,\dots,K$. At each iteration, the orchestrator executes the flow
% $
% \mathcal{D}_t \rightarrow \{\mathcal{D}_t^{(i)}\}_{i=1}^N \rightarrow \mathcal{D}_{t+1}$,
% where agents generate and evaluate parallel candidate designs $\{\mathcal{D}_t^{(i)}\}_{i=1}^N$ based on feedback, and the best-performing one is selected for the next iteration.

% Specifically, given the evaluated candidates $\{\mathcal{D}_t^{(i)}\}_{i=1}^N$, the next design is selected as
% \begin{equation}
% \mathcal{D}_{t+1} = \arg\min_{\mathcal{D}_t^{(i)}, \;\text{SEC}_i = 1} f(\text{WNS}_i, \text{TNS}_i, \text{Area}_i),
% \end{equation}
% ensuring that only functionally correct candidates are promoted.

% In addition, the orchestrator maintains iteration-level records, including candidate performance and transformation context, which are used to support subsequent strategy learning.

\subsubsection{Timing Analysis Agent.}

The timing analysis agent identifies timing bottlenecks through fine-grained slack analysis, but does not directly propose fixes. Rather than relying on coarse design-level feedback such as a single WNS value, it uses detailed post-synthesis timing reports to localize critical-path issues back to RTL structure.

Given the current design and timing report, the agent selects the top-$k$ critical paths by slack. For each path, it performs path-to-RTL mapping by locating the startpoint and endpoint registers and tracing the intermediate combinational logic back to the corresponding RTL regions. It then diagnoses likely root causes of delay, such as excessive combinational depth, high fanout, control--data coupling, or reconvergent logic, to guide subsequent RTL optimization agent.

% and assigns each path a likely optimization mode, such as combinational or latency-preserving sequential transformation.

% \subsubsection{Timing Analysis Agent.}

% The timing analysis agent identifies timing bottlenecks through fine-grained slack analysis, without directly proposing fixes. Unlike prior LLM-based RTL optimization methods that rely on coarse design-level feedback (e.g., a single WNS value), our approach leverages detailed timing information to precisely localize critical-path issues.

% Given the current design and its timing report, the agent selects the top-k critical paths based on slack distribution. For each path, it performs \emph{path-to-RTL mapping} by locating the startpoint and endpoint registers and tracing the intermediate combinational logic back to the corresponding RTL regions.

% The agent then diagnoses potential root causes of delay, such as excessive combinational depth, high fanout, control--data coupling, or reconvergent logic. Each path is also labeled by likely optimization mode, such as combinational restructuring or latency-preserving sequential refinement.

\subsubsection{RTL Optimization Agent.}

The RTL optimization agent generates $N$ candidate designs $\{\mathcal{D}_t^{(i)}\}_{i=1}^N$ in parallel based on timing analysis and the accumulated skill library $\mathcal{S}$ (detailed in~\Cref{sec:method2}). Guided by critical-path information and root-cause analysis, it applies equivalent transformations to handle true bottlenecks. It first queries $\mathcal{S}$ for pattern--strategy entries matching the diagnosed bottleneck. If a match is found, the corresponding skill-guided transformation is applied to $\mathcal{D}_t$. Otherwise, the LLM proposes a new transformation conditioned on the timing feedback.

This design balances \emph{exploration} and \emph{exploitation}. Exploration comes from generating multiple candidates in parallel, either by applying different transformations to the same bottleneck or by targeting different bottlenecks in the current iteration. Exploitation comes from promoting the best SEC-pass candidate $\mathcal{D}_{t+1}$ to the next iteration and from reusing effective skills in $\mathcal{S}$ across designs.

Compared with prior methods driven by coarse design-level metrics and fixed basic rewriting rules, our optimizer uses critical-path-level feedback and learned skills to guide adaptive RTL transformations, improving both search efficiency and optimization quality.\looseness=-1

\subsubsection{Evaluation Agent.}

The evaluation agent executes the synthesis and verification toolchain. Given a candidate design, it runs the commercial synthesis flow and sequential equivalence checking (SEC), and returns the resulting metrics
$(\text{WNS}_i, \text{TNS}_i, \text{Area}_i, \text{SEC}_i)$.

The evaluation agent performs no reasoning or report interpretation, ensuring a strict separation between execution and decision-making. This design guarantees that all optimization decisions are based solely on verified EDA feedback, avoiding hallucinations.

% \subsubsection{Skill Learning Agent.}
% In addition to the agents performing the RTL optimization, we also employ an auxiliary agent for skill learning, online.
% It reads the past optimization trajectories and 
% extract reusable knowledge. Within each iteration, it compares the parallel candidate designs based on their relative performance to identify effective transformations.

% Skill extraction is based on relative performance among SEC-passing candidates within each iteration, as detailed in the next section.
% Over time, this process enables Dr. RTL to accumulate optimization knowledge and guide future RTL modifications.

% The detailed group-relative learning mechanism is described in the next section.

\subsection{Group-Relative Skill Learning from Hierarchical Optimization Trajectories}\label{sec:method2}
A central goal of Dr.~RTL is not only to optimize the current design, but also to accumulate reusable optimization knowledge over time. Without such skill learning, each iteration would rely mainly on local trial-and-error, making it difficult to systematically reuse successful transformations across iterations and designs. A reusable skill library is therefore important for improving PPA more efficiently, reducing invalid or unhelpful transformations, and accelerating convergence through cross-design reuse, and providing interpretable knowledge of what works and why.

A key challenge, however, is determining which transformations are truly worth storing as skills. Because EDA outcomes are heuristic and design-dependent, absolute PPA values are often noisy and not directly comparable across candidates or iterations. This makes it difficult to attribute improvement to any single transformation, especially when multiple edits are composed over time.

\begin{figure}[!t]
  \centering
  % \vspace{-.2in}
  \includegraphics[width=1\linewidth]{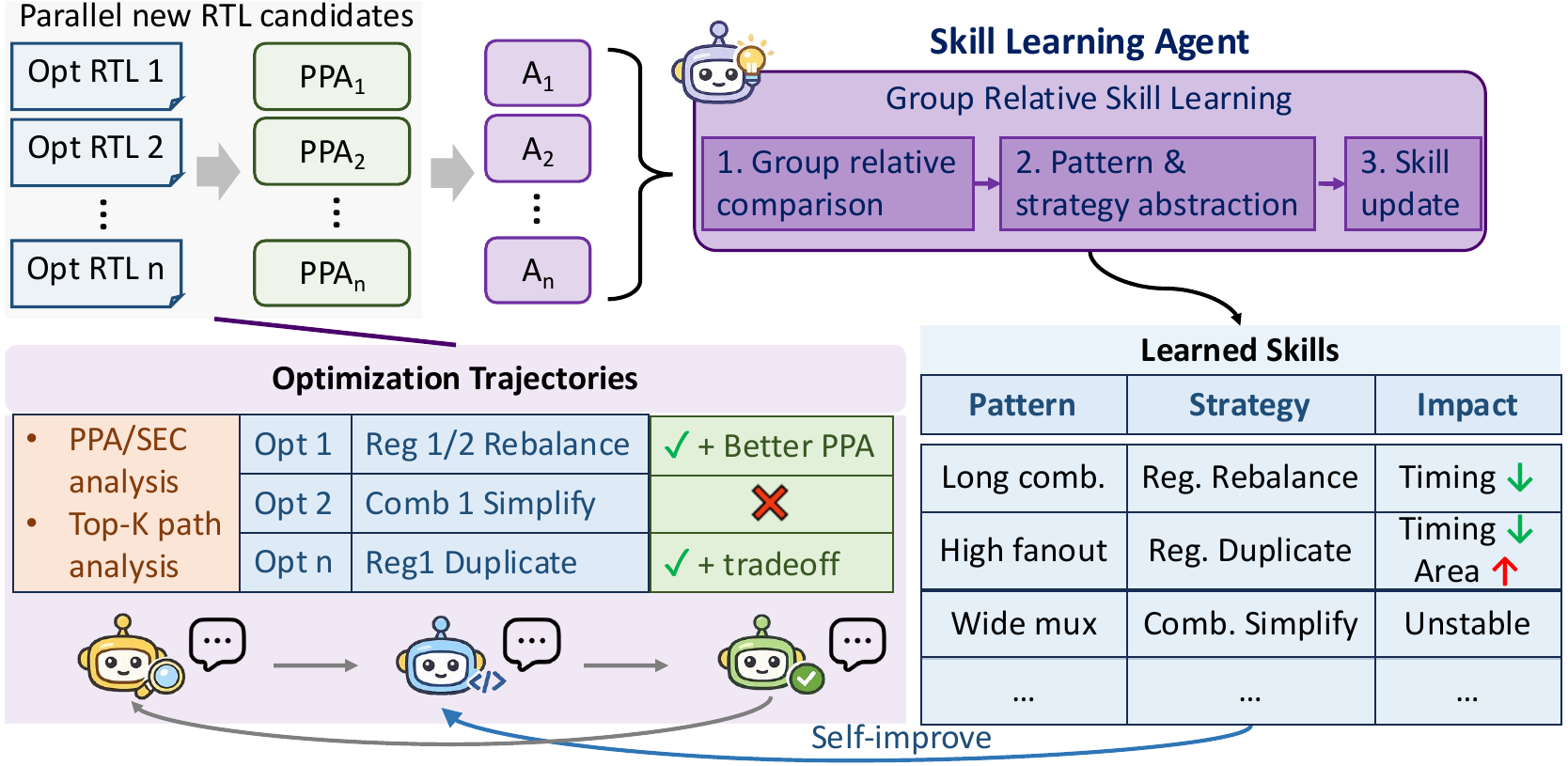}
  \vspace{-.3in}
  \caption{Group-relative skill learning. Parallel RTL candidates are compared to extract patterns and strategies, which are stored as reusable skills for continuous improvement.} 
  \label{fig:skill}
  \vspace{-.15in}
\end{figure}

To address this, we propose \emph{group-relative skill learning}, as illustrated in~\Cref{fig:skill}. The core idea is to evaluate a strategy not by its absolute PPA outcome, but by how well it performs relative to other candidates generated from the same parent RTL under the same timing feedback and tool settings. We refer to the set of parallel candidates $\{\mathcal{D}_t^{(i)}\}_{i=1}^N$ in iteration $t$ as a \emph{group}. This within-group comparison provides a more stable signal for identifying effective strategies without external supervision.

\textbf{Hierarchical optimization trajectory.}
Each candidate design is associated with an optimization trajectory that records its full optimization context, including the timing analysis, applied RTL transformations, and resulting evaluation outcomes. We organize these trajectories as a three-layer hierarchy. The top layer contains the iteration round, recording the temporal evolution of optimization. Within each round, the second layer stores all parallel candidate designs together with their PPA and SEC results, preserving the within-group context needed for relative comparison. At the lowest layer, it records critical-path-level information, including the analyzed bottlenecks, their structural patterns and root causes, the transformations applied, and the resulting outcomes. This hierarchical trajectory connects analysis, optimization actions, and evaluation results across multiple granularities, providing the structured context needed for comparison and skill extraction.\looseness=-1

\textbf{Group-relative comparison.}
Built on this trajectory representation, Dr.~RTL compares parallel candidates within each group by computing a relative advantage signal:
\begin{equation}
A_i = \frac{\textit{score}_i - \mu_t}{\sigma_t},
\label{eq:advantage}
\end{equation}
where $\mu_t$ and $\sigma_t$ denote the mean and standard deviation of the candidate scores (Eq.~\eqref{eq:score}) in iteration $t$. Intuitively, $A_i$ measures whether a candidate performs better or worse than its peers under matched conditions, highlighting strategies that consistently outperform alternatives in the same optimization context.

\textbf{Pattern--strategy abstraction.}
Guided by the relative advantage signal, we employ an auxiliary skill learning agent to analyze the trajectories and extract reusable pattern--strategy pairs as optimization skills. A pattern captures a recurring structural bottleneck on critical paths, such as deep FSM/decode logic, high-fanout control signals, wide comparisons, or mux-heavy selection logic. A strategy describes the corresponding transformation principle, such as condition pre-computation, signal replication, or selective register insertion. We will provide detailed case studies in~\Cref{expr:case}.

\textbf{Skill update.}
Based on these abstractions, Dr.~RTL updates a confidence-aware skill library by merging newly extracted pattern--strategy pairs with existing entries and accumulating simple empirical statistics, including occurrence count, SEC-pass count, and mean relative advantage across iterations and designs. Skills that repeatedly appear in successful trajectories are assigned higher confidence, while unreliable or ineffective ones are deprioritized as invalid strategies.

Overall, group-relative skill learning converts noisy absolute PPA feedback into more reliable relative signals, enabling continuous self-improvement without supervision or reward design.

%% file: section/4-experiments.tex
% \vspace{-.1in}
\section{Experimental Results} \label{sec:expr}
\subsection{Experimental Setup}

\begin{table*}[!t]
\vspace{-.4in}
\caption{PPA improvements of Dr. RTL over commercial logic synthesis on real-world RTL designs.}
\vspace{-.15in}
\resizebox{1\textwidth}{!}{ 

\begin{tabular}{c||cccc|ccc|ccc|c} \toprule
\multicolumn{1}{c||}{\textbf{Design}}     
& \multicolumn{4}{c|}{\textbf{Statistics}}                   
& \multicolumn{3}{c|}{\textbf{Commercial Synthesis}}                                                                                                                             
& \multicolumn{4}{c}{\cellcolor{agentrtl}\textbf{w/ Dr. RTL Optimization}}                                                                                                                                                                                       \\
\multicolumn{1}{c||}{}                                     
& \small{LoC} & \small{NoM} & \small{\#. Gate} & \small{\#. Reg}      
& \small{WNS ($ns$)} & \small{TNS ($ns$)} & \small{Area ($um^2$)} 
& \small{WNS ($ns$)} & \small{TNS ($ns$)} & \small{Area ($um^2$)} & \small{SEC Pass} \\ 
\midrule \midrule

vending   & 128 & 1 & 20272 & 4   & -0.27 & -1.02 & 20488 & -0.09 (-66.7\%) & -0.5 (-51.0\%)  & 20533 (0.2\%)  & 77\% \\
ticket    & 134 & 1 & 56    & 6   & -0.23 & -1.24 & 78    & -0.09 (-60.9\%) & -0.47 (-62.1\%) & 45 (-41.8\%)   & 65\% \\
lstm      & 135 & 4 & 8379  & 0   & -6.51 & -166.76 & 14828 & -4.86 (-25.3\%) & -116.16 (-30.3\%) & 4753 (-67.9\%) & 83\% \\
dsp       & 165 & 5 & 3345  & 196 & -2.61 & -154.64 & 4755 & -2.59 (-0.8\%)  & -147.42 (-4.7\%) & 4751 (-0.1\%)  & 83\% \\
communicate & 225 & 3 & 1023 & 232 & -0.4  & -73.08 & 2092 & -0.26 (-35.0\%) & -58.84 (-19.5\%) & 2446 (17.0\%)  & 95\% \\
spi1      & 332 & 2 & 647   & 131 & -0.32 & -33.42 & 1208 & -0.29 (-9.4\%)  & -33.53 (0.3\%)  & 1276 (5.7\%)   & 63\% \\
cpu\_fsm  & 354 & 1 & 13232 & 4163 & -0.82 & -429.73 & 32268 & -0.61 (-25.6\%) & -427.28 (-0.6\%) & 32157 (-0.3\%) & 96\% \\
aes       & 374 & 2 & 24294 & 1419 & -0.7  & -913.49 & 33755 & -0.67 (-4.3\%) & -876.71 (-4.0\%) & 33975 (0.7\%)  & 81\% \\
fifo      & 390 & 7 & 13740 & 4208 & -0.54 & -2002.1 & 36061 & -0.43 (-20.4\%) & -1681.97 (-16.0\%) & 36310 (0.7\%) & 96\% \\
spi2      & 441 & 3 & 856   & 292 & -0.26 & -28.19 & 1748 & -0.25 (-3.8\%)  & -27.75 (-1.6\%) & 1826 (4.5\%)   & 100\% \\
uart      & 447 & 4 & 851   & 135 & -0.38 & -23.17 & 1272 & -0.34 (-10.5\%) & -23.08 (-0.4\%) & 1107 (-13.0\%) & 87\% \\
controller& 528 & 1 & 213   & 8   & -0.38 & -2.85  & 235  & -0.33 (-13.2\%) & -2.57 (-9.8\%)  & 277 (18.0\%)   & 94\% \\
router    & 571 & 5 & 3036  & 609 & -0.53 & -289.91 & 5479 & -0.46 (-13.2\%) & -246.66 (-14.9\%) & 5575 (1.8\%) & 92\% \\
cpu\_pipe & 850 & 5 & 2845  & 364 & -0.38 & -23.24 & 2622 & -0.11 (-71.1\%) & -2.69 (-88.4\%) & 2313 (-11.8\%) & 90\% \\
pcie      & 923 & 7 & 1773  & 97  & -0.79 & -23.83 & 2156 & -0.44 (-44.3\%) & -19.75 (-17.1\%) & 1426 (-33.9\%) & 100\% \\
datapth   & 1065 & 7 & 6985 & 881 & -0.88 & -513.42 & 12137 & -0.87 (-1.1\%) & -504 (-1.8\%) & 12137 (0.0\%) & 92\% \\
i2c       & 1036 & 3 & 723  & 128 & -0.36 & -26.67 & 1290 & -0.35 (-2.8\%) & -25.96 (-2.7\%) & 1275 (-1.1\%) & 98\% \\
tv80      & 4615 & 5 & 4431 & 359 & -1.31 & -381.2 & 6044 & -1.22 (-6.9\%) & -362.09 (-5.0\%) & 6200 (2.6\%) & 92\% \\
arm\_cpu1 & 2070 & 1 & 11772 & 1222 & -5.24 & -3148.68 & 22172 & -5.19 (-1.0\%) & -3117.41 (-1.0\%) & 22257 (0.4\%) & 56\% \\
arm\_cpu2 & 1450 & 1 & 6132 & 736 & -1.01 & -662.7 & 10688 & -0.92 (-8.9\%) & -619.13 (-6.6\%) & 10995 (2.9\%) & 87\% \\

\midrule \midrule
\textbf{Avg.} 
& \textbf{812} & \textbf{3} & \textbf{6230} & \textbf{760} 
& \multicolumn{3}{c|}{\textbf{/}} 
& \cellcolor{agentrtl}\textbf{-21.3\%} 
& \cellcolor{agentrtl}\textbf{-16.9\%} 
& \cellcolor{agentrtl}\textbf{-5.8\%} 
& \cellcolor{agentrtl}\textbf{86\%} \\

\bottomrule                                              
\end{tabular}
}
\label{tbl:result}
\vspace{-.1in}
\end{table*}

\textbf{Diverse RTL design dataset.} We evaluate Dr.~RTL on 20 human-written Verilog designs selected using three criteria: complex functionality, full synthesizability, and size ranging from hundreds to thousands of lines of code. The dataset spans diverse IPs and systems, including buffers, processors, signal-processing modules, cryptographic designs, and SoCs, with a wide range of scales and structural complexity. Detailed statistics are summarized in~\Cref{tbl:result}, with an average of 812 LOC (min 128, max 4615) and 3 modules per design. We also report synthesized gate and register counts to reflect design complexity. For comparison, we further evaluate on all large designs from the representative benchmark of~\cite{yao2024rtlrewriter}, detailed in~\Cref{sec:baseline}.\looseness=-1

\begin{table}[!t]
% \vspace{-.3in}
\caption{Comparison with existing single-shot LLMs and LLM-based RTL optimization methods.\looseness=-1}%applied in optimization.}
\vspace{-.15in}
\resizebox{0.45\textwidth}{!}{ 

\begin{tabular}{c|c||ccc} \toprule
\multicolumn{2}{c||}{\textbf{Method}}                                                                  & WNS            & TNS            & Area          \\ \midrule
\multirow{2}{*}{\begin{tabular}[c]{@{}c@{}}Single-Shot\\      LLM\end{tabular}}   & Claude Opus      & -2.4\%           & -2.8\%           & -0.7\%          \\
                                                                                  & GPT 5.3          & -1.2\%           & 0.3\%            & 0.6\%           \\ \midrule
\multirow{2}{*}{\begin{tabular}[c]{@{}c@{}}SOTA   \\      Baselines$^\star$\end{tabular}} & RTLRewriter~\cite{yao2024rtlrewriter}      & -4.9\%           & -6.3\%           & -3.1\%          \\
                                                                                  & SymRTLo~\cite{wang2025symrtlo}          & -7.1\%           & -5.7\%           & -1.4\%          \\ \midrule
\rowcolor{agentrtl}\textbf{Ours}                                                                     & \textbf{Dr. RTL} & \textbf{-21.3\%} & \textbf{-16.9\%} & \textbf{-5.8\%} \\ \bottomrule
\end{tabular}

}
\begin{tablenotes}\footnotesize
\item $^\star$ We use the same backbone LLM, Claude Opus, for both the state-of-the-art LLM-based RTL optimization baselines and our Dr. RTL to ensure a fair comparison.
\end{tablenotes} 
\label{tbl:baseline2}
\vspace{-.3in}
\end{table}

\textbf{Evaluation workflow.}
We evaluate Dr.~RTL and all baselines under a unified protocol. All designs are synthesized with Synopsys Design Compiler using the Nangate 45\,nm library~\cite{NanGate} under fixed synthesis constraints for fair comparison. In particular, we use a tight clock period of 0.1\,ns to force aggressive synthesis optimization across all paths. Functional equivalence between design versions is verified by sequential equivalence checking using Cadence Jasper SEC~\cite{SEC}.\looseness=-1

For each design, every method runs 10 iterations with 5 parallel candidates per iteration, for 50 optimization attempts in total. At each iteration, the best SEC-passing candidate is selected by PPA and promoted to the next round. We report the best PPA improvement, SEC pass rate, and optimization trajectories. To explicitly evaluate skill generalization, we use 4-fold cross-validation in the skill application setting: in each fold, optimization skills are extracted from 15 designs and then applied to 5 unseen designs. The skill library is reset for each fold, and all prompts and examples are kept strictly fold-isolated, preventing leakage from benchmark overlap. This ensures that the reported results reflect cross-design skill reuse rather than design-specific memorization.\looseness=-1

\textbf{Implementation of Dr.~RTL.}
Dr.~RTL is currently implemented on the Claude Code command-line interface (CLI), while the framework is agent-CLI agnostic and can be readily adapted to other agentic coding environments through the same markdown-based workflow specification. Unless otherwise noted, we use Claude Opus for the main experiments, and evaluate other Claude models such as Sonnet and Haiku in the scaling study to cover different capability--cost trade-offs~\cite{claude}. We also include OpenAI GPT-5.3 and Qwen Coder 8B as baselines.

\subsection{Optimization Results}
\textbf{Per-design results.}
\Cref{tbl:result} summarizes Dr.~RTL’s optimization results across all 20 designs. Notably, Dr.~RTL improves timing on all 20 designs, achieving average WNS and TNS improvements of 21.3\% and 16.9\%, respectively, while even reducing area by 5.8\%. This suggests that the gains mainly come from structurally efficient RTL transformations rather than area-expensive trade-offs. In particular, 9/20 designs achieve Pareto improvements, 8/20 incur only minor area overhead ($<5\%$), and only 3/20 show noticeable area increase.

Dr.~RTL also maintains high reliability throughout optimization, achieving an average SEC pass rate of 86\%. Overall, these results show that Dr.~RTL delivers consistent timing improvement on realistic human-written RTL through iterative, feedback-driven optimization beyond advanced synthesis.

\textbf{Comparison with baseline methods.}
As shown in~\Cref{tbl:baseline2}, we compare Dr.~RTL with single-shot LLMs and representative prior iterative methods~\cite{yao2024rtlrewriter, wang2025symrtlo} on our real-world dataset. Single-shot models are evaluated in one pass, while iterative baselines use the same iteration budget and base model (i.e., Claude Opus) as Dr.~RTL. In contrast, prior iterative methods rely mainly on design-level feedback and pre-defined rules, and do not accumulate reusable knowledge from interaction with the synthesis environment.

Dr.~RTL consistently outperforms all baselines, achieving the best overall results in WNS, TNS, and area. Among the single-shot baselines, Claude Opus performs better than GPT-5.3, so we adopt it as the backbone model of Dr.~RTL. Notably, the iterative baselines~\cite{yao2024rtlrewriter, wang2025symrtlo} also use the same backbone LLM, indicating that the gains mainly come from our agentic optimization framework rather than the underlying model alone.

\begin{table}[!t]
% \vspace{-.35in}
\caption{Existing benchmarks fail to capture real RTL optimization. Dr.~RTL not only repairs manually degraded RTL, but also consistently improves the original human-written RTL beyond the benchmark's assumed upper bound.}
\vspace{-.15in}
\resizebox{0.51\textwidth}{!}{

\begin{tabular}{c|ccc|ccc|ccc} \toprule
                                  & \multicolumn{3}{c|}{\cellcolor[HTML]{FBE2D5}\textbf{\begin{tabular}[c]{@{}c@{}}Manual-Degraded \\      Suboptimal RTL\end{tabular}}}                                                                                                                                                       & \multicolumn{3}{c|}{\textbf{\begin{tabular}[c]{@{}c@{}}Human-Written \\      RTL (Baseline)\end{tabular}}} & \multicolumn{3}{c}{\cellcolor{agentrtl}\textbf{\begin{tabular}[c]{@{}c@{}}Better Optimized   \\   RTL    by Dr. RTL\end{tabular}}}                                                                                                                                                  \\ \cmidrule{2-10}

\multirow{-2}{*}{\textbf{Design}} & \small{WNS}                                                                                  & \small{TNS}                                                                                  & \small{Area}                                                                                & \small{WNS}                                                                                  & \small{TNS}                                                                                  & \small{Area}           & \small{WNS}                                                                                  & \small{TNS}                                                                                  & \small{Area}                                                                                 \\ \midrule \midrule
FFT& {\color[HTML]{C00000} \begin{tabular}[c]{@{}c@{}}-3.2\\      (+2\%)\end{tabular}}    & {\color[HTML]{C00000} \begin{tabular}[c]{@{}c@{}}-4446\\      (+0.4\%)\end{tabular}} & {\color[HTML]{4EA72E} \begin{tabular}[c]{@{}c@{}}160688\\      (-6\%)\end{tabular}} & -3.14         & -4431         & 170620         & {\color[HTML]{4EA72E} \begin{tabular}[c]{@{}c@{}}-1.73\\      (-45\%)\end{tabular}} & {\color[HTML]{4EA72E} \begin{tabular}[c]{@{}c@{}}-1746\\      (-61\%)\end{tabular}} & {\color[HTML]{4EA72E} \begin{tabular}[c]{@{}c@{}}110546\\      (-35\%)\end{tabular}} \\ \midrule
Huffman& {\color[HTML]{4EA72E} \begin{tabular}[c]{@{}c@{}}-2.23\\      (0\%)\end{tabular}}    & {\color[HTML]{C00000} \begin{tabular}[c]{@{}c@{}}-2705\\      (+0.4\%)\end{tabular}} & {\color[HTML]{4EA72E} \begin{tabular}[c]{@{}c@{}}73807\\      (-8\%)\end{tabular}}  & -2.23         & -2695         & 80009          & {\color[HTML]{4EA72E} \begin{tabular}[c]{@{}c@{}}-2.21\\      (-1\%)\end{tabular}}  & {\color[HTML]{4EA72E} \begin{tabular}[c]{@{}c@{}}-2562\\      (-5\%)\end{tabular}}  & {\color[HTML]{4EA72E} \begin{tabular}[c]{@{}c@{}}71223\\      (-11\%)\end{tabular}}  \\ \midrule
VMachine& {\color[HTML]{C00000} \begin{tabular}[c]{@{}c@{}}-0.27\\      (+145\%)\end{tabular}} & {\color[HTML]{C00000} \begin{tabular}[c]{@{}c@{}}-1.02\\      (+15\%)\end{tabular}}  & {\color[HTML]{C00000} \begin{tabular}[c]{@{}c@{}}20967\\      (+2\%)\end{tabular}}  & -0.11         & -0.89         & 20488          & {\color[HTML]{4EA72E} \begin{tabular}[c]{@{}c@{}}-0.08\\      (-27\%)\end{tabular}} & {\color[HTML]{4EA72E} \begin{tabular}[c]{@{}c@{}}-0.38\\      (-57\%)\end{tabular}} & {\color[HTML]{C00000} \begin{tabular}[c]{@{}c@{}}21989\\      (+7\%)\end{tabular}}   \\ \midrule
CNN& \multicolumn{9}{c}{{\color[HTML]{C00000} }}                                                                                                                                                                                                                                                                                                                                                                                                                                                                                                                                           \\ 
CPU& \multicolumn{9}{c}{\multirow{-2}{*}{{\color[HTML]{C00000} Cannot be synthesized using DC}}}    \\ \bottomrule                         
\end{tabular}

}
\label{tbl:baseline1}
\vspace{-.2in}
\end{table}

\begin{figure*}[!t]
  \centering
  \vspace{-.4in}
  \includegraphics[width=1\linewidth]{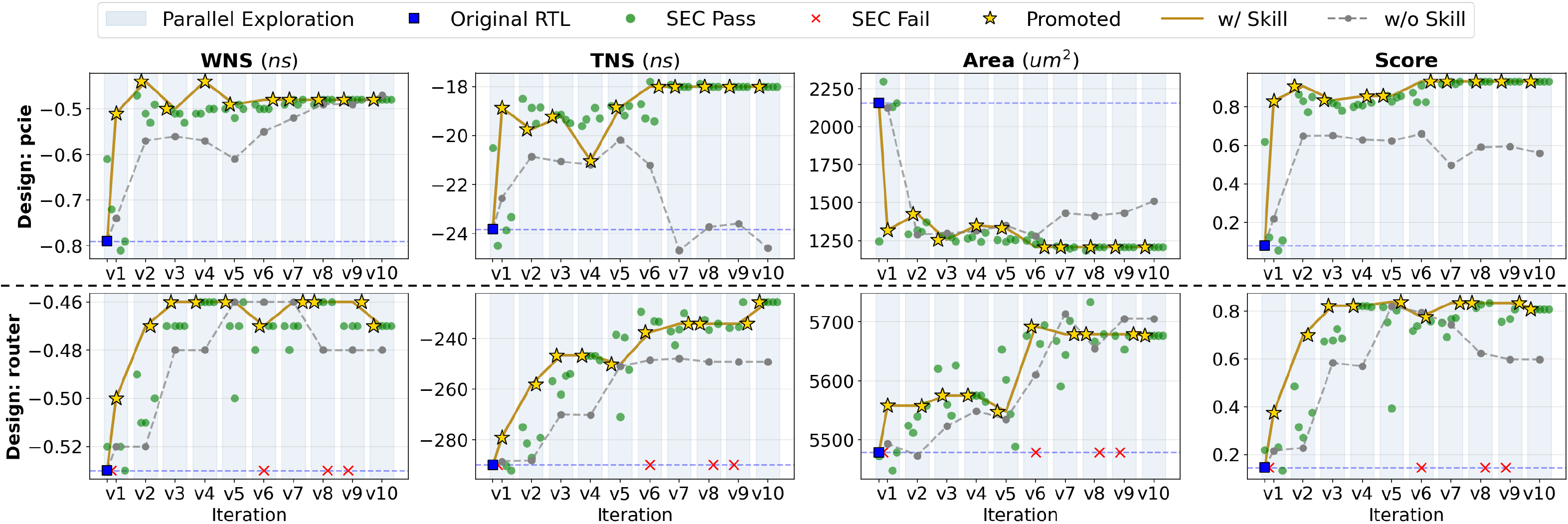}
  \vspace{-.25in}
  \caption{Optimization trajectories of Dr.~RTL across iterations of two design examples (router and pcie). WNS, TNS, area, and score improve steadily while preserving functional equivalence. Parallel exploration selects better candidates at each step, and the skill-enhanced variant achieves faster convergence and better PPA than the baseline without skills.} 
  \label{fig:trajectory}
  \vspace{-.1in}
\end{figure*}

\begin{figure}[!t]
  \centering
  % \vspace{-.35in}
  \includegraphics[width=1\linewidth]{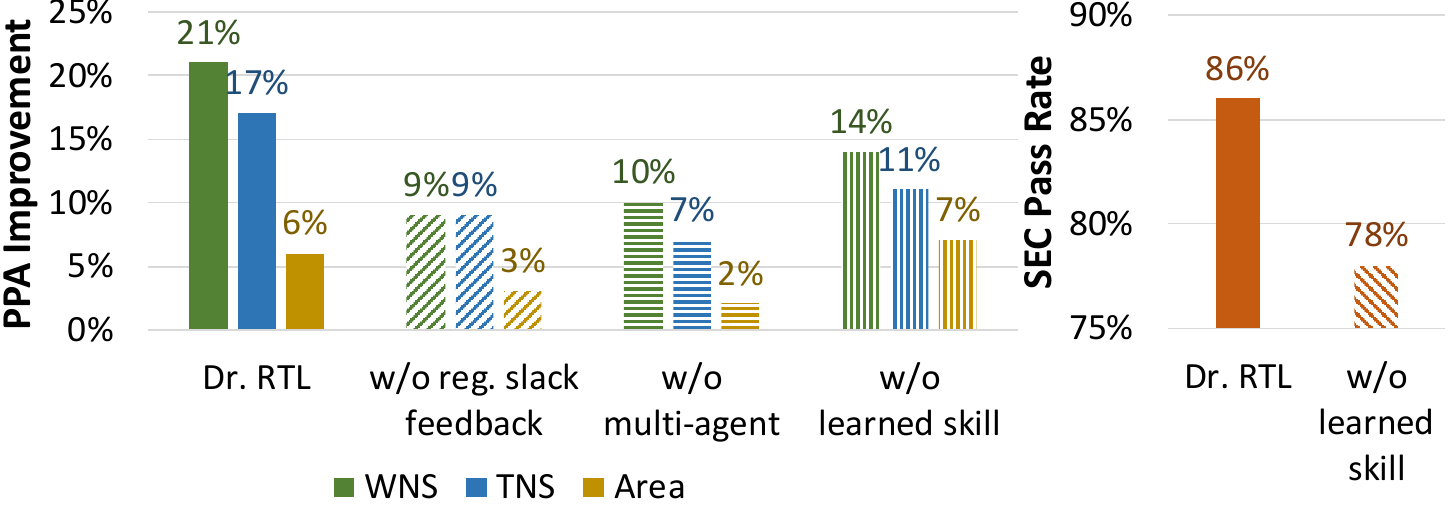}
  \vspace{-.2in}
  \caption{Ablation studies of Dr.~RTL on register-level slack feedback, multi-agent design, and learned skill library.}
  \label{fig:ablation}
  \vspace{-.15in}
\end{figure}

\subsection{Existing Benchmarks Miss Real Optimization}\label{sec:baseline}

We further evaluate Dr.~RTL on a representative RTL optimization benchmark~\cite{yao2024rtlrewriter}, which contains large designs with manually degraded RTL. We include all large designs provided by the benchmark, and note that two of them are not even synthesizable under commercial tools due to syntax issues, further exposing the gap between open-source and industrial EDA evaluation.\looseness=-1

We also apply Dr.~RTL to this benchmark, as shown in~\Cref{tbl:baseline1}. Dr.~RTL even further improves the original human-written RTL, which existing benchmarks treat as the optimization upper bound. In contrast, prior methods start from manually degraded designs and aim only to recover the original RTL. This shows that the original human-written RTL is not a true upper bound and can still be further optimized under realistic industrial evaluation. More broadly, these results suggest that realistic RTL optimization should be evaluated by whether a method can improve strong human-written RTL beyond synthesis, rather than merely recover it from artificial degradation.

% \textbf{Improvement over existing RTL optimization benchmarks.}
% We first evaluate Dr.~RTL on a representative RTL optimization benchmark~\cite{yao2024rtlrewriter}, which consists of large designs with manually degraded or suboptimal RTL. We adopt all the large designs provided in this benchmark. Note that these designs were originally evaluated using Yosys, and two designs do not fully support commercial synthesis flows due to syntax errors.

% As shown in \Cref{tbl:baseline1}, Dr.~RTL not only repairs degraded designs but also consistently improves timing beyond the original human-written RTL baselines, which are typically treated as the upper bound in prior work. This demonstrates that Dr.~RTL can uncover non-trivial optimization opportunities even when starting from already optimized RTL.

\subsection{Optimization Trajectory Visualization}
We visualize the optimization trajectories of Dr.~RTL in~\Cref{fig:trajectory}. At each iteration, multiple RTL candidates are explored in parallel, and the best SEC-passing design is promoted to the next round. Dr.~RTL shows steady improvement in WNS and TNS across iterations, indicating effective convergence. Compared with the variant without the learned skill library, the skill-enhanced version converges faster and reaches better PPA, showing that learned optimization knowledge improves both quality and convergence efficiency.\looseness=-1

\subsection{Ablation Studies}\label{sec:abl}

We conduct ablation studies on the key components of Dr.~RTL, as shown in~\Cref{fig:ablation}. Removing register-level slack feedback causes the largest drop, reducing WNS/TNS improvement from 21\%/17\% to 9\%/9\%, highlighting the importance of fine-grained timing information for understanding real critical paths and their root causes. Replacing the multi-agent framework with a single-agent setup reduces performance to 10\%/7\%, confirming that role decomposition helps structure the optimization process and enables more effective exploration. Disabling the skill library lowers WNS/TNS improvement to 14\%/11\% and reduces the SEC pass rate from 86\% to 78\%, indicating that reusable optimization knowledge not only improves search quality but also steers the agent away from ineffective or risky transformations.

\begin{figure*}[!t]
  \centering
  \vspace{-.4in}
  \includegraphics[width=1\linewidth]{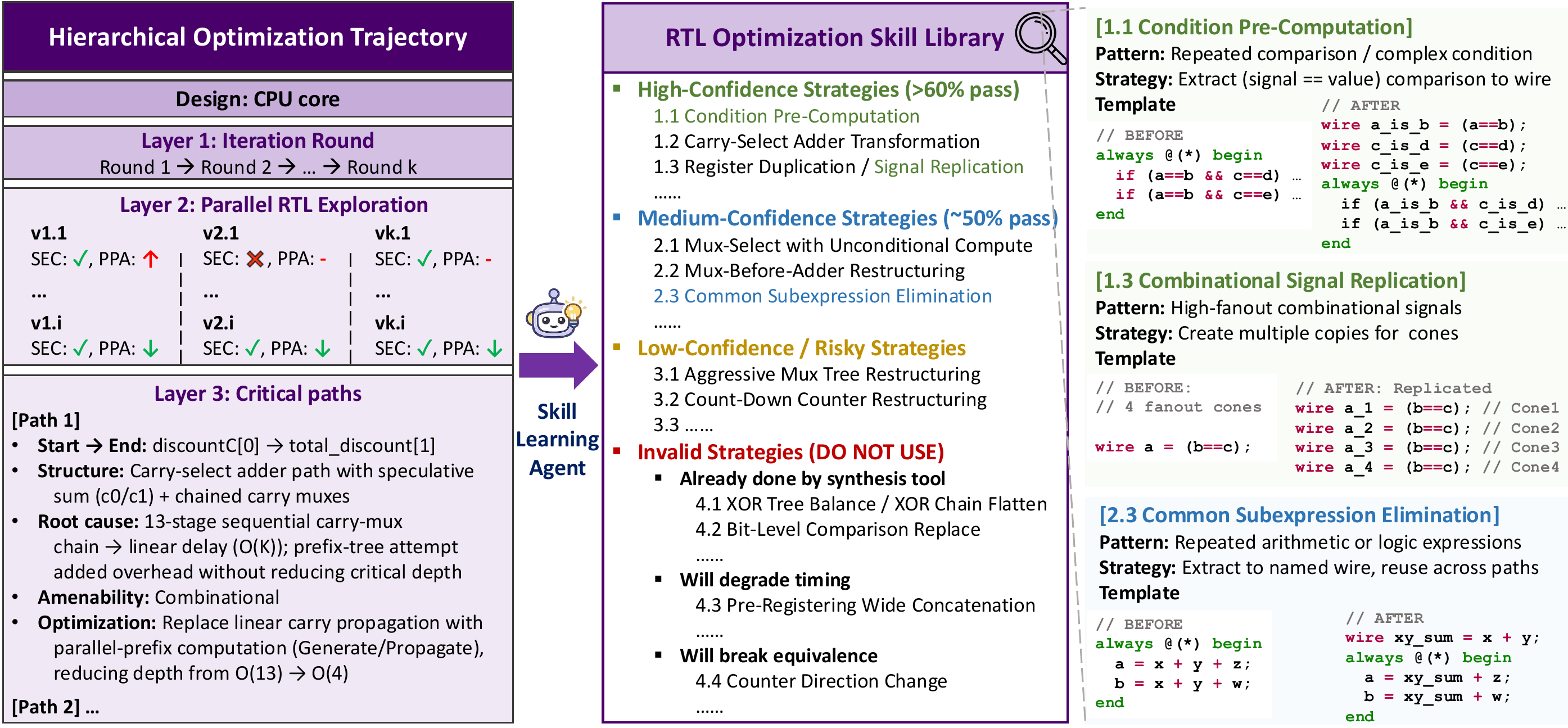}
  \vspace{-.25in}
  \caption{Distilling hierarchical trajectory into skill library. Optimization trajectories are organized across iterations, parallel explorations, and critical paths, enabling discovery of reusable RTL optimization strategies with different confidence levels and associated transformation templates.} 
  \label{fig:skill-demo}
  \vspace{-.1in}
\end{figure*}

\subsection{Runtime and Cost Analysis}
The runtime of Dr.~RTL is dominated by EDA execution, since each candidate requires synthesis and sequential equivalence checking. Because the $N$ candidates in each iteration are evaluated in parallel, wall-clock runtime scales mainly with the number of iterations rather than the total number of candidates, i.e., $\text{runtime} \approx K \times T_{\text{EDA}}$, where $K=10$ in our setup and $T_{\text{EDA}}$ is the runtime of one parallel EDA evaluation round. In practice, synthesis dominates $T_{\text{EDA}}$, ranging from minutes to hours depending on design size. Compared with traditional manual RTL optimization, this parallel evaluation enables a substantially more efficient workflow that can run continuously (24$\times$7). In our setup, optimizing all 20 designs required roughly one week of wall-clock time and about \$50 in LLM usage. Overall, Dr.~RTL provides a favorable cost--performance trade-off while substantially reducing manual effort.

% The runtime of Dr.~RTL is dominated by EDA tool execution, as each candidate requires synthesis and sequential equivalence checking (SEC). A key advantage of Dr.~RTL is that the $N$ candidate designs in each iteration are evaluated in parallel, so the wall-clock runtime scales mainly with the number of iterations instead of the total number of candidates. As a result, the overall runtime can be approximated as $\text{runtime} \approx K \times T_{\text{EDA}}$, where $K=10$ in our setup and $T_{\text{EDA}}$ denotes the runtime of one parallel EDA evaluation round. In practice, synthesis dominates $T_{\text{EDA}}$, ranging from minutes to hours depending on design size.

% Compared with traditional manual RTL optimization, this parallel evaluation yields a substantially more efficient workflow that can run continuously. In our setup, optimizing all 20 designs required roughly one week of wall-clock time and about \$50 in LLM usage. Because LLM cost scales roughly with RTL code length and remains negligible relative to EDA runtime, Dr.~RTL offers a favorable cost--performance trade-off while substantially reducing manual effort.

\begin{figure}[!t]
  \centering
  % \vspace{-.35in}
  \includegraphics[width=0.78\linewidth]{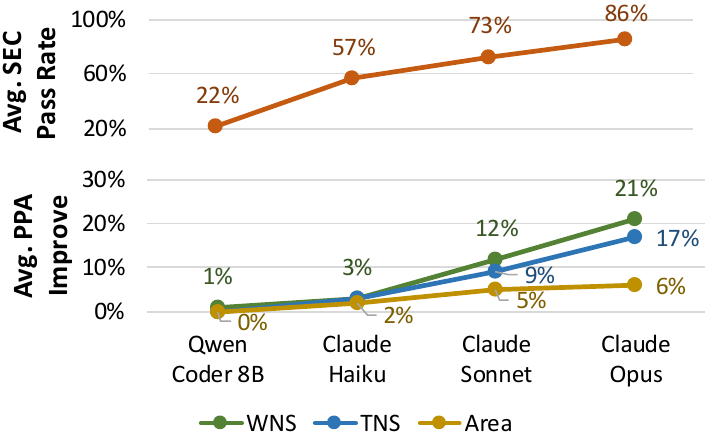}
  % \vspace{-.1in}
  \caption{Dr. RTL scales with LLM capability.} 
  \label{fig:scale}
  \vspace{-.1in}
\end{figure}

%% file: section/5-discussion.tex
\section{Discussion} \label{sec:disc}

\subsection{Skill Discovery and Interpretability}\label{expr:case}

\Cref{fig:skill-demo} shows how Dr.~RTL converts raw optimization trajectories into reusable RTL optimization knowledge. Hierarchical trajectories record iteration rounds, parallel exploration, and critical-path analyses with root-cause diagnoses, providing structured logs for skill extraction. In our current implementation, this process yields 47 entries in total: 12 high-confidence strategies, 16 medium-confidence strategies, 6 low-confidence strategies, and 13 avoid strategies.\looseness=-1

The discovered skills are organized by confidence level according to their empirical success rates under real EDA feedback. High-confidence skills correspond to consistently effective transformations (e.g., logic simplification, fanout management), while medium-confidence skills represent conditionally useful strategies (e.g., restructuring or resource sharing), and low-confidence skills correspond to more aggressive or design-dependent transformations. Dr. RTL also explicitly identifies avoiding strategies that are ineffective, absorbed by synthesis, or violate equivalence.\looseness=-1

Each skill is externalized as a pattern--strategy pair with an implementation template, enabling direct reuse. This confidence-aware organization allows Dr.~RTL to prioritize reliable transformations while adaptively exploring more complex strategies when beneficial. Overall, Dr.~RTL converts raw trajectory experience into structured, interpretable, and reusable knowledge, enabling systematic and adaptive agentic RTL optimization.

\subsection{Dr. RTL Scales with LLM Capability}
We study how Dr.~RTL scales with different LLM backbones, as shown in~\Cref{fig:scale}. Stronger models consistently improve both optimization quality and reliability. Average WNS/TNS improvement increases from 1\%/0\% with Qwen Coder 8B, to 3\%/2\% with Claude Haiku, to 12\%/9\% with Claude Sonnet, and to 21\%/17\% with Claude Opus. Meanwhile, the average SEC pass rate rises from 22\% to 57\%, 73\%, and 86\%. These results suggest that Dr.~RTL scales effectively with backbone capability. As smaller models continue to improve, fine-tuned or domain-specialized LLMs are a promising direction for better privacy, customization, and deployment flexibility.

\subsection{Impact of EDA Tools}
\begin{figure}[!t]
  \centering
  % \vspace{-.35in}
  \includegraphics[width=1\linewidth]{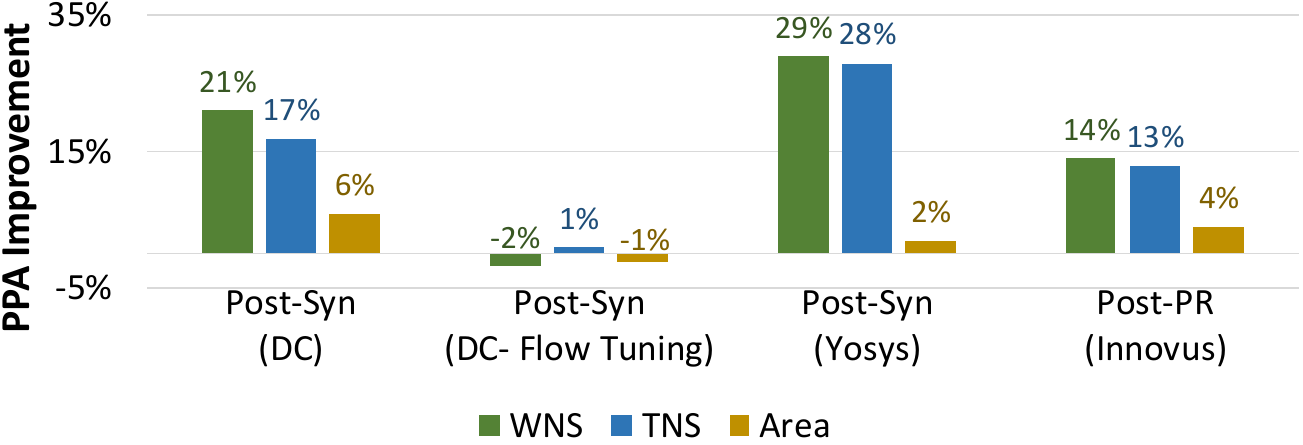}
  \vspace{-.25in}
  \caption{Impact of different EDA tools on Dr.~RTL results.\looseness=-1} 
  \label{fig:tool}
  \vspace{-.1in}
\end{figure}

As shown in~\Cref{fig:tool}, we evaluate Dr.~RTL across multiple EDA tools and design stages, to distinguish true RTL-level optimization from gains that might depend on a particular backend tool or flow setting.
Dr.~RTL consistently improves timing across all settings, from open-source synthesis to commercial synthesis and post-route evaluation. The gains are largest with Yosys, where the weaker synthesis leaves more room for RTL improvement, while they are smaller but still clear with commercial DC and after place-and-route in Innovus, where timing has already been more aggressively optimized. In contrast, simple DC flow tuning yields only marginal benefit, suggesting that parameter tuning is limited, while substantial further gains still come from improving the RTL itself.
Overall, these results show that Dr.~RTL effectively generalizes across both tools and design stages.

%% file: section/6-conclusion.tex
\section{Conclusion}\label{sec:concl}

We present Dr.~RTL, an agentic framework for RTL timing optimization that formulates the task as an iterative, closed-loop process through interaction with industrial EDA tools, achieving consistent timing improvement with minimal area overhead on real-world designs. Beyond performance, our work advocates a shift toward agentic design automation driven by autonomous reasoning, accumulated chip design knowledge, and EDA environment interaction. By coupling realistic evaluation with reusable skill learning, Dr.~RTL takes a step toward practical agentic design automation, and paves the way for future systems built on smaller specialized models, richer exploration strategies, and human-in-the-loop refinement.